\documentclass[runningheads]{llncs}
\usepackage[T1]{fontenc}
\usepackage{cite}
\usepackage{url}
\usepackage{enumitem}
\usepackage{microtype}
\usepackage{caption}
\usepackage{amsmath,amssymb,amsfonts}
\usepackage{algorithmic}
\usepackage{graphicx}
\usepackage{subfigure}
\usepackage{textcomp}
\usepackage{xcolor}
\usepackage{float}
\usepackage{ragged2e}
\usepackage{marvosym}
\usepackage{booktabs, makecell, multirow, tabularx}
\usepackage{color}
\usepackage{hyperref}
\hypersetup{
    colorlinks=true,
    linkcolor=blue,
    filecolor=blue,
    urlcolor=blue,
    citecolor=blue,
}
\newcommand{\hash}{{\ttfamily \#}}
\begin{document}
\title{PCLMix: Weakly Supervised Medical Image Segmentation via Pixel-Level Contrastive Learning and Dynamic Mix Augmentation}
\titlerunning{PCLMix: Weakly Supervised Medical Image Segmentation}
%
\author{Yu Lei \and Haolun Luo \and Lituan Wang\textsuperscript{(\Letter)} \and Zhenwei Zhang \and Lei Zhang}
\authorrunning{Y. Lei et al.}
%
\institute{College of Artificial Intelligence, Sichuan University, Chengdu, China\\
\email{lituanwang@scu.edu.cn}}
\maketitle              
\begin{abstract}
In weakly supervised medical image segmentation, the absence of structural priors and the discreteness of class feature distribution present a challenge, i.e., how to accurately propagate supervision signals from local to global regions without excessively spreading them to other irrelevant regions? To address this, we propose a novel weakly supervised medical image segmentation framework named PCLMix, comprising dynamic mix augmentation, pixel-level contrastive learning, and consistency regularization strategies. Specifically, PCLMix is built upon a heterogeneous dual-decoder backbone, addressing the absence of structural priors through a strategy of dynamic mix augmentation during training. To handle the discrete distribution of class features, PCLMix incorporates pixel-level contrastive learning based on prediction uncertainty, effectively enhancing the model's ability to differentiate inter-class pixel differences and intra-class consistency. Furthermore, to reinforce segmentation consistency and robustness, PCLMix employs an auxiliary decoder for dual consistency regularization. In the inference phase, the auxiliary decoder will be dropped and no computation complexity is increased. Extensive experiments on the ACDC dataset demonstrate that PCLMix appropriately propagates local supervision signals to the global scale, further narrowing the gap between weakly supervised and fully supervised segmentation methods. Our code is available at \href{https://github.com/Torpedo2648/PCLMix}{https://github.com/Torpedo2648/PCLMix}. 

\end{abstract}
\section{Introduction}
\label{sec_intro}
In medical image segmentation, although fully supervised learning methods have demonstrated excellent performance, they rely on acquiring extensive and high-quality densely annotated data, which is undeniably time-consuming and costly. To overcome this reliance, recent research has turned to the utilization of weakly labeled and unlabeled data for model training, employing methods like semi-supervised learning (SSL) \cite{Mittal_Tatarchenko_Brox_2021, Souly_Spampinato_Shah_2017, Tarvainen_Valpola_2017} and weakly supervised learning (WSL) \cite{Khoreva_Benenson_Hosang_Hein_Schiele_2017, Pathak_Krahenbuhl_Darrell_2015, Wei_Liang_Chen_Shen_Cheng_Feng_Zhao_Yan_2017}. However, SSL often requires precise annotation of partial images in the dataset \cite{Zhang_Zhuang}. As an attractive alternative, WSL, which only requires sparse annotations for effective model training, has garnered widespread attention.

Weak annotations, including image-level labels, sparse annotations, and noisy annotations, are more readily available in practice \cite{xu2015learning}. Among these forms, scribble stands out as one of the most convenient weak labels, i.e., our focused scenario in this work, demonstrating substantial potential in medical image segmentation \cite{Lin_Dai_Jia_He_Sun_2016}. Yet, in scribble-supervised segmentation, the absence of structural priors and the discreteness of class feature distribution introduce a new challenge: how to accurately propagate the supervision signals from local to global without excessively spreading them to other irrelevant regions?

\begin{figure}
\centering
\includegraphics[width=0.5\textwidth]{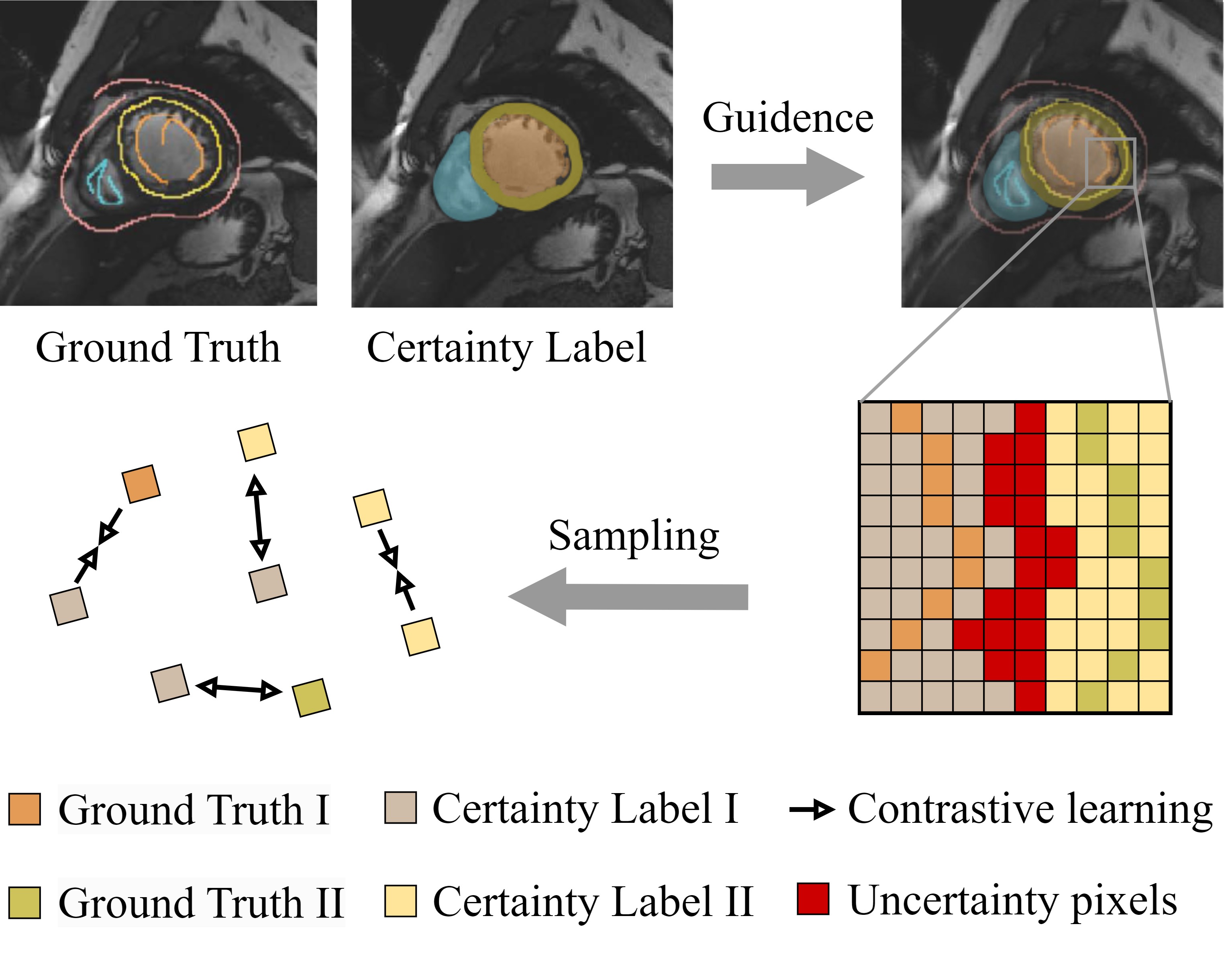}
\caption{The core idea of uncertainty-guided pixel-level contrastive learning for scribble-supervised medical image segmentation.}
\label{ugpcl}
\end{figure}

To address this question, significant progress has been made in previous studies \cite{Zheng_Hayashi_Oda_Kitasaka_Mori, Zhou_Xu_Zhou_Tong, Luo_Hu_Liao_Zhai_Song_Wang_Zhang, Grandvalet_Bengio_2004}. The main idea revolves around spreading the supervision signal of labels from sparse local regions to the entire image by generating pseudo labels. Recent notable works in this direction include DLPMS \cite{Luo_Hu_Liao_Zhai_Song_Wang_Zhang} and TriMix \cite{Zheng_Hayashi_Oda_Kitasaka_Mori}. The former proposes a dual-branch network strategy to generate pseudo labels, alleviating the sparsity issue of annotated labels to some extent. The latter employs joint training of three independent networks along with mix augmentation, further narrowing the gap between scribble-supervised methods and fully supervised methods. However, these approaches treat the classification of each pixel independently, overlooking the internal correlations among pixels.

One method to enhance internal correlations among pixels is pixel-level contrastive learning. Recently, Wang et al. \cite{Wang_Lu_Lai_Wen_Kong} proposed uncertainty-guided pixel-level contrastive learning for semi-supervised medical image segmentation, reducing noise sampling in the contrastive sample construction process through uncertainty maps. An intuitive idea is to apply uncertainty-guided pixel-level contrastive learning to weakly supervised medical image segmentation. However, in practice, this remains an unresolved challenge: How to accurately 
define similar pixels given sparse annotations? Additionally, while existing methods may involve model diversity, they do not fully exploit the inherent structural diversity within different models, inevitably leading to the loss of information from various perspectives.

In this article, we present PCLMix, a robust framework for weakly supervised medical image segmentation designed to address these challenges. PCLMix is structured upon a heterogeneous dual-decoder backbone, incorporating mix augmentation, pixel-level contrastive learning, and dual consistency regularization. For mix augmentation, PCLMix dynamically shuffles image-label pairs within the same batch and mixes them from two different views to generate new samples. These samples are then fed into the network for further training. Subsequently, as illustrated in Figure \ref{ugpcl} PCLMix utilizes the outputs of the dual branches to obtain pseudo labels and estimate uncertainty. It strategically selects pixels with high certainty and those within scribble annotations as anchors for pixel-level contrastive learning, thereby reducing noise sampling through uncertainty guidance. Furthermore, we introduce two regularization terms to impose heterogeneous consistency constraints and mixed consistency constraints on model training. To streamline the inference process, both decoders are engaged in training, while during the inference stage, only the main branch (CNN decoder) is utilized for predictions. Our evaluations on the public cardiac dataset ACDC demonstrate the promising performance of PCLMix, achieving comparable or even superior segmentation accuracy when compared to fully supervised approaches. The main contributions of our work are summarized as follows:
\begin{itemize}[label=\textbullet]
    \item We design a dynamic mix augmentation strategy tailored for heterogeneous dual-branch decoders, which enables PCLMix to adaptively mix different samples within the same batch during training, facilitating accurate propagation of supervision signals from local to global.
    \item To the best of our knowledge, We introduce uncertainty-guided pixel-level contrastive learning into the domain of weakly supervised medical image segmentation for the first time, which effectively strengthens the correlation between pixels, thereby preventing the excessive diffusion of supervision signals.
    \item We propose weakly supervised regularization from dual consistency constraints, encompassing both heterogeneous consistency and mixed consistency. They enhance the robustness and stability of the model by penalizing inconsistent segmentations between the main decoder and the auxiliary decoder.
\end{itemize}

\section{Related Work}
\label{sec_realted}

\subsection{Weakly Supervised Medical Image Segmentation}
To alleviate the time-consuming task of pixel-level segmentation annotation, numerous studies have explored the use of various forms of sparse annotation for semantic segmentation, such as scribble-level, bounding box-level, and image-level annotations \cite{Tajbakhsh_Jeyaseelan_Li_Chiang_Wu_Ding_2020}. In this paper, we focus on scribble, which serves as an intuitive and efficient alternative to dense annotation. Its primary challenge lies in the lack of supervision for object structure and uncertainty in boundary delineation. To address this, SribbleSup \cite{Lin_Dai_Jia_He_Sun_2016} employed a graph-based approach to propagate information to unlabeled pixels. ScribbleAlone \cite{Can_Chaitanya_Mustafa_Koch_Konukoglu_Baumgartner_2018} proposed to train a CNN using scribbles and then refine the CNN predictions using Conditional Random Fields (CRFs). Additionally, RLoss \cite{Tang_Perazzi_Djelouah_Ayed_Schroers_Boykov_2018} introduced a CRF-based regularization term that can be optimized alongside the segmentation loss function, rather than being used for post-processing with CRFs.

In recent years, Scribble2Label architecture \cite{Lee_Jeong_2020} suggested generating certainty pseudo labels by combining pseudo-labeling and label filtering under prediction consistency constraints. MAAG \cite{valvano2021learning} proposed incorporating an attention-gated mechanism in adversarial training to generate segmentation masks, enhancing target localization across various resolutions. DMPLS \cite{Luo_Hu_Liao_Zhai_Song_Wang_Zhang} introduced a dual-CNN-branch network, which produces high-quality pseudo labels by combining the outputs from two branches at random. Although these methods reduce annotation costs, there is still a significant performance gap compared to training with dense annotations, limiting their applicability and feasibility in clinical practice. 

\subsection{Mix Augmentation}
Data augmentation is very important for mitigating overfitting with finite training data and improving the generalization capacity of neural networks. Mix augmentation, an important data augmentation method, involves combining two images along with their corresponding labels. In contrast to traditional augmentation methods (such as rotation and flipping), this approach can provide more information for sparse annotations through mixing operations. Despite the potentially less realistic appearance of mixed images, they contribute to model training \cite{chaitanya2019semi}.

Until now, many mix augmentation methods have been proposed. MixUp \cite{Zhang_Cisse_Dauphin_Lopez-Paz} suggests constructing a new sample pair by taking the weighted average of the features and labels of two existing sample pairs. ManifoldMixUp \cite{Verma_Lamb_Beckham_Najafi_Mitliagkas_Lopez-Paz_Bengio_2019} extends the mixing operations to hidden features of input images. Cutout \cite{Devries_Taylor} randomly removes a square region from an image, while CutMix \cite{Yun_Han_Chun_Oh_Yoo_Choe_2019} replaces the removed region with the corresponding region from another image. Puzzle Mix \cite{Kim_Choo_Song_2020} introduces a new mixing method based on saliency and local statistics. Co-mix \cite{Kim_Choo_Jeong_Song_2021} broadens the mixing operation from two images to encompass multiple images, achieving significant success in enhancing the diversity of mixed images' supermodules. TriMix \cite{Zheng_Hayashi_Oda_Kitasaka_Mori} incorporates mix augmentation strategies into triple networks with the same structure, imposing consistency regularization under more rigorous perturbations. However, these methods do not fully exploit the diversity of model structures, inevitably losing complementary information from different perspectives.

\subsection{Contrastive Learning}
In image-level representation learning, contrastive learning can effectively leverage unlabeled data to learn meaningful visual representations. Its core idea is to enhance the discriminative power of learned visual representations by narrowing the similarity of positive pairs (similar samples) and separating dissimilar pairs (negative samples) under certain similarity constraints. The key challenge in image-level contrastive learning lies in how to construct contrastive samples. He \cite{He_Fan_Wu_Xie_Girshick_2020} proposed a viable solution by introducing a memory bank and momentum contrast, thereby increasing the number of contrastive samples.

In the context of semi-supervised image segmentation, some works \cite{codella2018skin, wang2021exploring, Zheng_Yang_2021} have extended contrastive learning from the image-level to the pixel-level. The main idea is to construct sample pairs directly for labeled data and use pseudo labels or spatial structures for unlabeled data. Moreover, to alleviate the potential issue of noise sampling during the construction of sample pairs, Wang \cite{Wang_Lu_Lai_Wen_Kong} further suggested using prediction uncertainty to guide sampling. However, in weakly supervised image segmentation, limited by the sparsity of annotations, how to construct rich and effective pixel-level contrastive samples while minimizing noise sampling remains a pressing issue to be addressed.

\subsection{Consistency Regularization}
The consistency regularization strategy is based on the premise that if the same image undergoes perturbations, the results should be consistent. Consistency constraints find wide applications in the fields of image translation and semi-supervised learning. CycleGAN \cite{Zhu_Park_Isola_Efros_Winter_Gogh_Monet_Photos} enhances the image-to-image translation capability by incorporating forward and backward consistency. In the context of semi-supervised learning, stability predictions for unlabeled images are obtained by imposing consistency on two augmented versions of the input image \cite{Tarvainen_Valpola_2017, Laine_Aila_2016, Ouali_Hudelot_Tami_2020}. CycleMix \cite{Zhang_Zhuang} proposes leveraging consistency at both global and local levels to exploit the properties of mixed invariance and the interdependence of segmentation structures. However, these methods also fail to fully exploit the diversity of model structures.

\section{Methodology}
\label{sec_method}

\begin{figure}[htbp]
\centering
\includegraphics[width=0.9\textwidth]{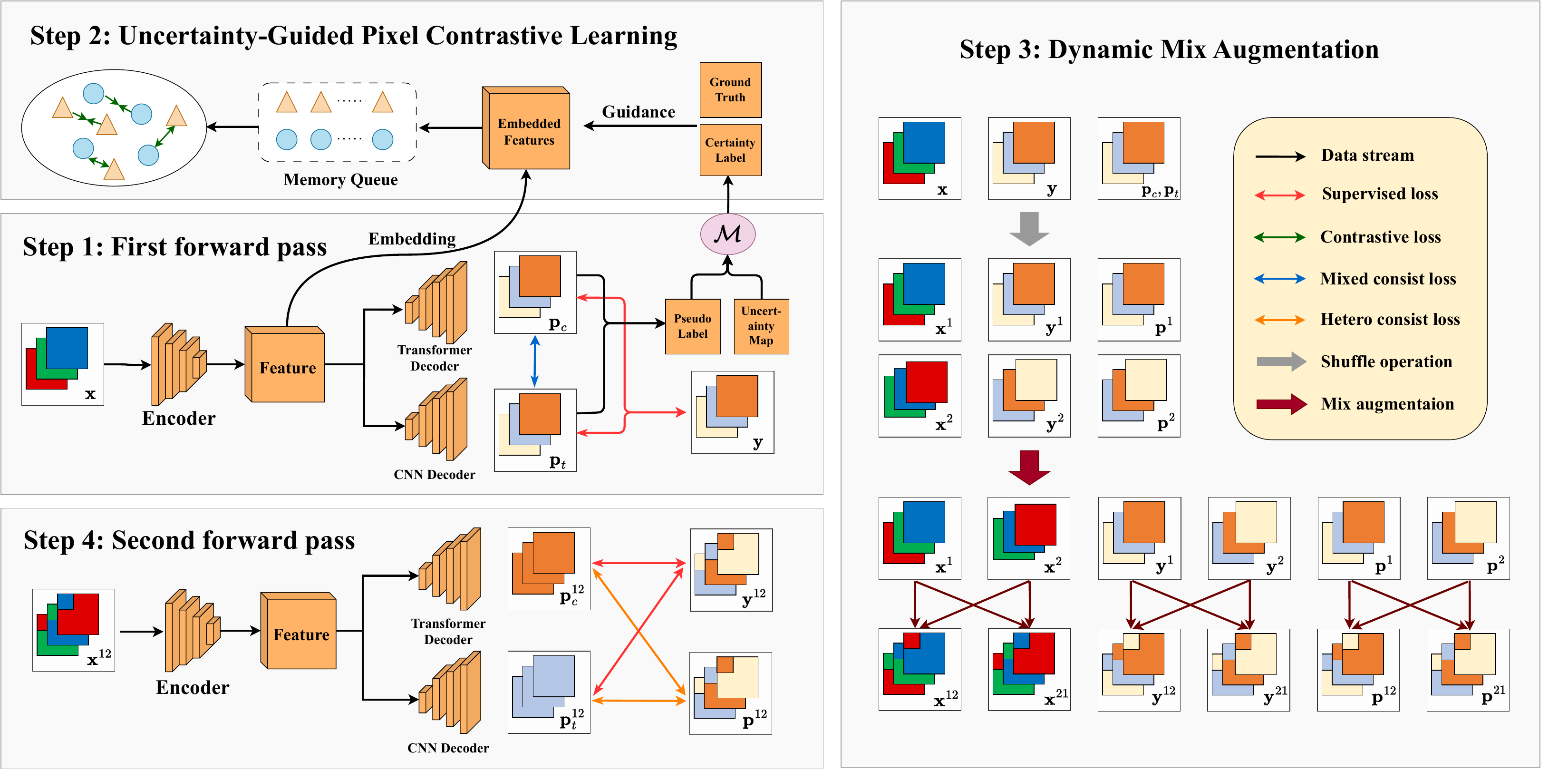}
\caption{The overview of PCLMix. PCLMix is built upon a heterogeneous dual-decoder backbone. It follows a four-step process: 1) data enters the network, generating pseudo labels and uncertainty maps. 2) Confirmed labels and scribbles construct a memory queue for contrastive learning. 3) The prediction segmentaions are shuffled and mixed with images and labels to create augmented data, 4) which is then fed back into the network. The entire training process will be driven by supervised loss, contrastive loss, and consistency loss.}
\label{pclmix}
\end{figure}

\subsection{Overview}
In scribble-supervised learning, each sample in the dataset $\textbf{D}$ includes an image denoted as $\textbf{X}$ and its corresponding scribble annotation denoted as $\textbf{Y}={(s_{1}, l_{1}), (s_{2}, l_{2}), \cdots, (s_{K}, l_{K})}$, where $K$ represents the number of scribble categories, $s_{k}$ refers to the scribble of category $k$, and $l_k$ represents the label of category $k$. Figure \ref{pclmix} illustrates the PCLMix architecture, which is based on a one-encoder dual-decoders design featuring a shared encoder $\mathcal{F}{e}$ and two independent decoders, denoted as $\mathcal{F}{c}$ (CNN-based) and $\mathcal{F}{t}$ (Transformer-based). While $\mathcal{F}{t}$ serves as an auxiliary decoder during network training, it is not used during the inference phase. This design allows PCLMix to leverage the structural differences between the decoders without increasing computational complexity.

PCLMix consists of three main components: dynamic mix augmentation, uncertainty-guided pixel-level contrastive learning, and dual consistency regularization. Dynamic mix augmentation expands the sample space by randomly mixing different samples within each batch, aiding in the propagation of sparse supervision signals from local to global regions in scribble annotations. However, mixed samples may have incomplete and discontinuous class feature distributions, potentially causing over-diffusion of supervision signals. To address this, uncertainty-guided pixel-level contrastive learning uses prediction uncertainty to construct contrastive samples at the pixel level, reducing noise sampling and improving intra-class consistency and inter-class distinctiveness. Finally, dual consistency regularization leverages the structural diversity of the heterogeneous dual-branch decoders, enhancing prediction stability by penalizing inconsistent segmentations between the main and auxiliary decoders. The overall optimization objective encompasses the scribble-supervised loss $\ell_{sup}$, contrastive learning loss $\ell_{ctr}$, and consistency regularization loss $\ell_{con}$, formulated as follows:
\begin{equation}
\mathcal{L}_{total}=\ell_{sup}+\lambda_{ctr}\ell_{ctr}+\lambda_{con}\ell_{con}
\end{equation}
where both of $\lambda_{ctr}$ and $\lambda_{con}$ are the balancing parameters. More details about different modules will be introduced in subsequent sections.

\subsection{Dynamic Mix Augmentation From Heterogeneous Views}
\label{subsec_mix_aug}
It has been proved that mix augmentation strategies can enrich the information available for scribble annotations through mixing operations\cite{Zhang_Zhuang}. However, current methods are often based on single-branch or multi-branch networks with identical architecture (CNN or Transformer), leading to the inevitable loss of complementary information from different perspectives. In contrast, the heterogeneous dual-branch architecture of PCLMix allows a mini-batch to generate two new mini-batches from different viewpoints during training.

It is assumed that, during each training iteration, a mini-batch $\{\textbf{x},\textbf{y}\}$ is obtained. And let $p_c$ and $p_t$ denote the segmentation predictions of $\mathcal{F}_{c}$ and $\mathcal{F}_{t}$. This forward propagation process can be formulated as:
\begin{equation}
(\textbf{p}_c, \textbf{p}_t)=\mathcal{F}(\textbf{x};\theta_{e},\theta_{c},\theta_{t})
\label{eq_2}
\end{equation}
where $\mathcal{F}(\cdot)$ refers to the heterogeneous dual-branch network. Following the classical mixing operation in previous works\cite{Zhang_Cisse_Dauphin_Lopez-Paz, Yun_Han_Chun_Oh_Yoo_Choe_2019}, we initially randomize $\{\textbf{x}, \textbf{y}\}$ in two distinct orders, resulting in two shuffled batches $\{\textbf{x}^1, \textbf{y}^1\}$ and $\{\textbf{x}^2, \textbf{y}^2\}$. Simultaneously, $(\textbf{p}_c, \textbf{p}_t)$ is also shuffled in the same order, yielding $(\textbf{p}^1, \textbf{p}^2)$.

Then, we apply mix augmentation to the two shuffled batches to generate two new batches $\{\textbf{x}^{12},\textbf{y}^{12}\}$ and $\{\textbf{x}^{21},\textbf{y}^{21}\}$, which differ from each other because the mix augmentation we employ is asymmetric. Similarly, $(\textbf{p}^1, \textbf{p}^2)$ undergoes mixing to obtain $\textbf{p}^{12}$ and $\textbf{p}^{21}$. Let $(\textbf{p}_c^{12},\textbf{p}_t^{12})=\mathcal{F}(\textbf{x}^{12};\theta_{e},\theta_{c},\theta_{t})$ and $(\textbf{p}_c^{21},\textbf{p}_t^{21})=\mathcal{F}(\textbf{x}^{21};\theta_{e},\theta_{c},\theta_{t})$ represent the predictions generated by the dual-branch network for the new batches. To measure the similarity between these predictions and the labels, we employ the partial cross-entropy loss \cite{Tang_Djelouah_Perazzi_Boykov_Schroers_2018}, which can be defined as:
\begin{equation}
    \mathcal{L}_{pce}(\textbf{m}_c,\textbf{m}_t,\textbf{n})=-\sum\limits_{j \in \textbf{J}} \sum\limits_{k \in K}((1-\lambda_{t}) \textbf{m}_c^{j, k}+\lambda_{t} \textbf{m}_t^{j, k}) \log \textbf{n}^{j, k}
\label{eq_3}
\end{equation}
where $\textbf{m}_c$ and $\textbf{m}_t$ are the prediction probabilities generated by $\mathcal{F}_{c}$ and $\mathcal{F}_{t}$ respectively, $\textbf{n}$ is the ground truth label, $\textbf{J}$ is the collection of pixels in scribbles, $K$ stands for the number of scribble categories. For each $j$-th pixel and $k$-th category, $\textbf{m}_c^{j, k}$ and $\textbf{m}_t^{j,k}$ indicate the predicted values in $\textbf{m}_c$ and $\textbf{m}_t$, and $\textbf{n}^{j, k}$ is the corresponding ground truth label. The hyperparameter $\lambda_{t}$ balances predictions between the primary (CNN) and auxiliary (Transformer) decoders.

Utilizing the loss function specified in Equation \ref{eq_3}, the supervision loss for dynamic mix augmentation includes two components: unmixed loss and mixed loss, which can be calculated as follows:
\begin{equation}
    \ell_{sup}=\underbrace{\mathcal{L}_{pce}(\textbf{p}_c,\textbf{p}_t,\textbf{y})}_{unmix}+ \underbrace{\mathcal{L}_{pce}(\textbf{p}_c^{12},\textbf{p}_t^{12},\textbf{y}^{12})+\mathcal{L}_{pce}(\textbf{p}_c^{21},\textbf{p}_t^{21},\textbf{y}^{21})}_{mixed}
\end{equation} 

\subsection{Uncertainty-Guided Pixel-Level Contrastive Learning}
The introduction of mix augmentation in scribble-supervised learning may heighten the challenge of handling incomplete and disjointed annotations, potentially leading to less accurate local predictions. To this end, we introduce a pixel-level contrastive learning mechanism tailored for scribble annotations. By encouraging high similarity among pixels of the same category and dissimilarity among pixels of different categories, the model learns to distinguish pixels from the semantic level. In pixel-level contrastive learning, accurately determining the true categories of pixels is crucial for anchor sampling. However, the categories of pixels in unlabeled regions remain unknown. Existing methods address this by using pseudo labels but may introduce noise as well \cite{Wang_Lu_Lai_Wen_Kong}. To mitigate this, the anchor sampling in PCLMix is guided by prediction uncertainty from the dual-branch network. Importantly, this method remains applicable when utilizing other forms of sparse annotations.

\subsubsection{Uncertainty Estimation.}
To quantitatively estimate the uncertainty of pseudo labels, we employ predictive entropy as an approximate metric. Based on our heterogeneous dual-branch decoders, we concurrently consider the entropy of the outputs from both decoders, and balance them with a hyperparameter $\lambda_{t}$. In our work, the definition of uncertainty for a specific pixel $j$ can be expressed by the following formula:
\begin{equation}
    \textbf{u}_j=-\textbf{q}_j\log(\textbf{q}_j+\epsilon)
\end{equation}
\begin{equation}
    \textbf{q}_j=(1-\lambda_{t}) \textbf{p}_c^j+\lambda_{t} \textbf{p}_t^j
\end{equation}
where $\textbf{q}_j$ represents the averaged prediction probabilities of $\mathcal{F}_{c}$ and $\mathcal{F}_{t}$ for pixel $j$ across different classes, and $\epsilon$ is a small constant to prevent floating-point overflow. We believe that pixels with higher entropy are deemed uncertain, and their pseudo labels are unreliable. Conversely, pixels with lower entropy will be incorporated into the certainty map as anchor points to be sampled for contrastive learning. The corresponding confirmed pseudo label for pixel $j$ can be calculated by:
\begin{equation}
\textbf{PL}_j=\left\{
\begin{array}{ll}
    \textbf{y}_j, & \text{if } j\in \textbf{J}\\
    -\textbf{1}, & \text{if } j\notin \textbf{J}\ \text{and}\ \textbf{u}_j\ge H\\
    \operatorname{argmax}({\textbf{q}_j}), & \text{if } j\notin \textbf{J}\ \text{and}\ \textbf{u}_j< H
\end{array}\right.
\end{equation}
where $\textbf{J}$ is the collection of pixels in the scribble annotation, $H$ is the threshold that distinguishes between high and low entropy, and $\textbf{y}_j$ is the ground truth label for pixel $j$. Specifically, for pixels within scribbles, their corresponding pseudo labels are assigned the class label of the scribble. For pixels in other regions, if their uncertainty exceeds the threshold $H$, the corresponding confirmed pseudo label will be set to $-\textbf{1}$, indicating that they are uncertain pixels and will not be utilized during contrastive learning.

\subsubsection{Pixel-Level Contrastive Learning.}
Inspired by \cite{Wang_Lu_Lai_Wen_Kong}, we utilize contrastive learning in a low-resolution feature space to improve computational efficiency and capture more extensive semantic information. To efficiently manage computational overhead, we establish a memory queue to store acquired samples, holding prototype vectors alongside their corresponding pixel categories. The widely used InfoNCE loss function for computing contrastive loss is employed. In each iteration, a random selection of $N$ anchor points is made, and the contrastive loss is calculated for each anchor. The overall contrastive loss is then determined by averaging the losses from all selected anchors. The calculation process is illustrated as follows:
\begin{gather}
 \ell_{ctr}^i=-\frac{1}{M_p} {\sum_{j=1}^{M_p}} \log \frac{e^{\cos \left(\textbf{v}_i, \textbf{v}_j^{+}\right) / \tau}}{e^{\cos \left(\textbf{v}_i, \textbf{v}_j^{+}\right) / \tau}+\sum_{k=1}^{M_n} e^{\cos \left(\textbf{v}_i, \textbf{v}_k^{-}\right) / \tau}} \nonumber\\
 \ell_{ctr}=\frac{1}{N}\sum_{i=1}^{N}{\ell_{ctr}^i}
\end{gather}
Here, the prototype vector $\textbf{v}_i$ corresponds to pixel $i$, and $M_p$ and $M_n$ represent the counts of positive and negative samples for the same pixel. $\textbf{v}_j^+$ and $\textbf{v}_k^-$ denote the positive ($j$-th) and negative ($k$-th) prototype vectors, respectively. The cosine similarity function is denoted by $\cos$, and $\tau$ acts as the temperature hyperparameter.

\subsection{Regularization of Supervision via Dual Consistency}
PCLMix employs dual consistency to impose regularization on the model, which comprises two aspects: heterogeneous consistency and mixed consistency. In practical implementation, they are incorporated into the overall loss function as a regularization term $\ell_{con}$.

\subsubsection{Heterogeneous Consistency.}
As mentioned in Section \ref{sec_intro}, the CNN branch serves as the primary predictor in fact, while the Transformer branch acts as the auxiliary predictor. This auxiliary effect is reflected in the consistency of prediction between heterogeneous dual-branch decoders during training. In our implementation, we utilize the Mean Squared Error (MSE) loss as the metric to quantify the inconsistency between the dual decoders:
\begin{equation}
    \mathcal{L}_{het}=\operatorname{MSE}(\textbf{p}_c,\textbf{p}_t)+\operatorname{MSE}(\textbf{p}_c^{12},\textbf{p}_t^{12})+\operatorname{MSE}(\textbf{p}_c^{21},\textbf{p}_t^{21})
\end{equation}
where $\textbf{p}_c$ and $\textbf{p}_t$ are segmentation predictions from $\mathcal{F}_{c}$ and $\mathcal{F}_{t}$ in Equation \ref{eq_2}.

\subsubsection{Mixed Consistency.}
Intuitively, a robust model should exhibit consistent predictions under various perturbations, such as mix augmentation. This suggests that the segmentation predictions for the blended image and the combined predictions of the segmentation should ideally remain highly consistent, which can be formulated as follows:
\begin{equation}
    \mathcal{M}(\mathcal{F}(\textbf{x}^1),\mathcal{F}(\textbf{x}^2))=\mathcal{F}(\mathcal{M}(\textbf{x}^1,\textbf{x}^2))
    \label{eq_10}
\end{equation}
where $\textbf{x}^1$ and $\textbf{x}^2$ are two different batches of images, $\mathcal{M}(\cdot,\cdot)$ is the mixing function, and $\mathcal{F}(\cdot)$ is the segmenter. To our delight, based on dynamic mix augmentation in Section \ref{subsec_mix_aug}, the Equation \ref{eq_10} can be reformulated as:
\begin{equation}
    \left\{\begin{array}{l} 
    \textbf{p}^{12} = (1-\lambda_{t})\cdot \textbf{p}_c^{12}+\lambda_{t}\cdot \textbf{p}_t^{12}\\
    \textbf{p}^{21} = (1-\lambda_{t})\cdot \textbf{p}_c^{21}+\lambda_{t}\cdot \textbf{p}_t^{21}\\
\end{array}\right.
\end{equation}
Here, $\textbf{p}^{12}$ and $\textbf{p}^{21}$ are respectively the mixed predictions of the new images $\textbf{x}^{12}$ and $\textbf{x}^{21}$. Following previous work \cite{Zhang_Zhuang, Chen_He_2021, Grill_Strub_Altché_Tallec_Richemond_Elena_Doersch_Pires_Guo_Azar_et_al._2020}, we define the mixed consistency loss as the negative cosine similarity between two segmentation results:
\begin{equation}
  \begin{aligned}
    \mathcal{L}_{mix} = \mathcal{L}_{ncs}(\textbf{p}^{12},\textbf{q}^{12})+\mathcal{L}_{ncs}(\textbf{p}^{21},\textbf{q}^{21})
  \end{aligned}
\end{equation}
where $\textbf{q}^{12}\triangleq (1-\lambda_{t})\cdot \textbf{p}_c^{12}+\lambda_{t}\cdot \textbf{p}_t^{12}$ and $\textbf{q}^{21} \triangleq (1-\lambda_{t})\cdot \textbf{p}_c^{21}+\lambda_{t}\cdot \textbf{p}_t^{21}$ represent the dual decoders' average predictions of the mixed images, and $\mathcal{L}_{ncs}(\cdot,\cdot)$ denotes the negative cosine similarity.

Now, the regularization term $\ell_{con}$ is obtained:
\begin{equation}
\ell_{con}=\mathcal{L}_{het}+\lambda_{mix}\mathcal{L}_{mix}
\end{equation}
where $\lambda_{mix}$ is a hyperparameter to balance different regularization components.

\section{Experiments}

\subsection{Experimental Setup}
\subsubsection{Datasets.} We conducted evaluation on the ACDC dataset \cite{Bernard_Lalande_Zotti_Cervenansky_Yang_Heng_Cetin_Lekadir_Camara_Gonzalez_Ballester_et_al._2018} through five-fold cross-validation. The dataset contains 200 2-dimensional cine-MRI scans from 100 patients and is publicly available. For each patient, manual annotations are provided for the right ventricle (RV), left ventricle (LV), and myocardium (MYO) at end-diastolic (ED) and end-systolic (ES) phases. And the scribble annotations proposed by Valvano et al. \cite{valvano2021learning} were utilized. 

\subsubsection{Metircs.} Considering the unique nature of medical image segmentation tasks, we chose the Dice coefficient (Dice Score) and 95\% Hausdorff Distance (HD95) as performance metrics for our model evaluation. The Dice Score emphasizes internal segmentation accuracy by measuring the ratio of intersecting areas between predicted and ground truth masks. On the other hand, HD95 highlights boundary delineation precision by considering the maximum distance between pixels in the predicted and ground truth masks.

\subsubsection{Implementation Details.} We use ResNet-50 as our shared encoder, the decoder part of U-Net and Swin-Transformer as our heterogeneous dual-decoder. We leverage CutMix for mix augmentation, specifying a cropped area ratio of $0.2$. The parameters $\lambda_{ctr}, \lambda_{con}$, and $\lambda_{mix}$, responsible for balancing various components within the overall objective, are set to $0.15, 1.0$, and $1.0$ respectively. The balancing parameter $\lambda_t$ governing the interplay between the dual-branch decoders is assigned a value of $0.4$. During network training, we normalize the intensity of each slice to the range $[0, 1]$ and resize them to $256\times 256$ pixels. SGD serves as our optimizer with a weight decay of $0.0005$ and a momentum of $0.9$. The learning rate starts at $0.03$ and follows a polynomial scheduler strategy, gradually reducing to $0.001$ throughout training. The implementation is conducted using the PyTorch library and the models are trained on two NVIDIA RTX 3090 GPUs, with a batch size of $12$ and a total of $60$ thousand iterations.

\begin{table}
\centering
\caption{Comparison of our method with state-of-the-art scribble-supervised approaches on the ACDC dataset. All evaluations are conducted through a \textit{5-fold cross-validation}. The average results (standard deviation) for other methods are sourced from \cite{Verma_Lamb_Beckham_Najafi_Mitliagkas_Lopez-Paz_Bengio_2019}. \textbf{Bold} indicates the best performance, while \underline{underline} signifies the second-best. $\uparrow$ denotes lower values are better, $\downarrow$ indicates higher values are preferred, and $^\dag$ marks methods employing an ensemble strategy.}
\resizebox{\textwidth}{!}{
\begin{tabular}{*{9}{c}}
\toprule
\multirow{2}*{Methods} & \multicolumn{2}{c}{RV} & \multicolumn{2}{c}{Myo} & \multicolumn{2}{c}{LV} & \multicolumn{2}{c}{Avg} \\
\cmidrule(lr){2-3}\cmidrule(lr){4-5}\cmidrule(lr){6-7} \cmidrule(lr){8-9}
& Dice$\uparrow$ & HD95$\downarrow$ & Dice$\uparrow$ & HD95$\downarrow$ & Dice$\uparrow$ & HD95$\downarrow$ & Dice$\uparrow$ & HD95$\downarrow$ \\
\midrule
upper bound & 88.2 (9.5) & 6.9 (10.8) & 88.3 (4.2) & 5.9 (15.2) & 93.0 (7.4) & 8.1 (20.9) & 89.8 (7.0) & 7.0 (15.6) \\
baseline \cite{Lin_Dai_Jia_He_Sun_2016} & 62.5 (16.0) & 187.2 (35.2) & 66.8 (9.5) & 165.1 (34.4) & 76.6 (15.6) & 167.7 (55.0) & 68.6 (13.7) & 173.3 (41.5) \\
\midrule
RW \cite{Grady_2006} & 81.3 (11.3) & 11.1 (17.3) & 70.8 (6.6) & 9.8 (8.9) & 84.4 (9.1) & 9.2 (13.0) & 78.8 (9.0) & 10.0 (13.1) \\
USTM \cite{Liu_Yuan_Gao_He_Wang_Tang_Tang_Shen_2022} & 81.5 (11.5) & 54.7 (65.7) & 75.6 (8.1) & 112.2 (54.1) & 78.5 (16.2) & 139.6 (57.7) & 78.6 (11.9) & 102.2 (59.2) \\
S2L \cite{Lee_Jeong_2020} & 83.3 (10.3) & 14.6 (30.9) & 80.6 (6.9) & 37.1 (49.4) & 85.6 (12.1) & 65.2 (65.1) & 83.2 (9.8) & 38.9 (48.5) \\
MLoss \cite{Kim_Ye_2020} & 80.9 (9.3) & 17.1 (30.8) & 83.2 (5.5) & 28.2 (43.2) & 87.6 (9.3) & 37.9 (59.6) & 83.9 (8.0) & 27.7 (44.5) \\
EM \cite{Grandvalet_Bengio_2004} & 83.9 (10.8) & 25.7 (44.5) & 81.2 (6.2) & 47.4 (50.6) & 88.7 (9.9) & 43.8 (57.6) & 84.6 (8.9) & 39.0 (50.9) \\
RLoss \cite{Tang_Perazzi_Djelouah_Ayed_Schroers_Boykov_2018} & 85.6 (10.1) & 7.9 (12.6) & 81.7 (5.4) & 6.0 (6.9) & 89.6 (8.6) & 7.0 (13.5) & 85.6 (8.0) & 6.9 (11.0) \\
DMPLS \cite{Luo_Hu_Liao_Zhai_Song_Wang_Zhang} & 86.1 (9.6) & 7.9 (12.5) & 84.2 (5.4) & 9.7 (23.2) & 91.3 (8.2) & 12.1 (27.2) & 87.2 (7.7) & 9.9 (21.0) \\
SC-Net \cite{Zhou_Xu_Zhou_Tong} & 86.2 (7.1) & \textbf{4.6 (3.8)} & 83.9 (8.8) & 6.7 (16.3) & 91.5 (8.3) & 8.1 (14.1) & 87.2 (6.3) & 6.5 (13.9) \\
TriMix \cite{Zheng_Hayashi_Oda_Kitasaka_Mori} & \underline{87.7 (2.8)} & 8.9 (4.6) & \textbf{86.4 (2.2)} & \underline{4.3 (1.6)} & \textbf{92.3 (3.0)} & \underline{4.4 (1.9)} & \textbf{88.8 (3.2)} & \underline{5.9 (3.5)} \\
\midrule
PCLMix (ours) & \textbf{88.2 (4.7)} & \underline{5.3 (7.1)} & \underline{85.6 (3.6)} & \textbf{3.3 (1.9)} & \underline{92.2 (3.3)} & \textbf{3.7 (2.8)} & \underline{88.7 (2.7)} & \textbf{4.1 (2.9)} \\
\bottomrule
\label{quantComp}
\end{tabular}}
\end{table}

\subsection{Quantitative Comparison}
\subsubsection{Compared Methods.} We compare our proposed PCLMix with recent state-of-the-art scribble-supervised methods, including RW \cite{Grady_2006} (utilizing random walker for additional labeling), USTM \cite{Liu_Yuan_Gao_He_Wang_Tang_Tang_Shen_2022} (uncertainty-aware mean teacher and transformation consistency model), S2L \cite{Lee_Jeong_2020} (scribble to label), MLoss \cite{Kim_Ye_2020} (employing Mumford-Shah loss), EM 
\cite{Grandvalet_Bengio_2004} (focused on entropy minimization), DMPLS \cite{Luo_Hu_Liao_Zhai_Song_Wang_Zhang} (dual-branch mixed supervision), SC-Net \cite{Zhou_Xu_Zhou_Tong} (superpixel-guided scribble walking and class-wise contrastive regularization), and TriMix \cite{Zheng_Hayashi_Oda_Kitasaka_Mori} (triple models and mix augmentation). Besides, baseline configuration employs a 2D U-Net trained with scribble annotations and utilizing the pCE loss \cite{Lin_Dai_Jia_He_Sun_2016}. Moreover, the upper bound accuracy was reached using fully dense annotations. To ensure fairness, all methods are implemented without any ensemble strategies, and evaluation results are based on 5-fold cross-validation.

\subsubsection{Results.} Table \ref{quantComp} presents the results of our quantitative comparative experiments on the ACDC dataset. Clearly, compared to existing methods, PCLMix demonstrates more prom-ising outcomes. In terms of improvement magnitude, PCLMix outperforms the baseline significantly on the ACDC dataset, increasing the Dice score by 20.1 and reducing the HD95 by 169.2. Furthermore, when compared with state-of-the-art methods (SOTAs), PCLMix achieves the second-best Dice performance and the best HD95 performance among all scribble-supervised approaches. This indicates that, in overall segmentation performance, PCLMix is on par with SOTAs while achieving substantial improvement in boundary segmentation. It’s worth noting that, compared to the upper limit, our method exhibits a slight decrease in Dice (only 1.1) but surprisingly good results in HD95, with a relative reduction of 41.4\%. This shows that our approach successfully overcomes the inherent con-straints associated with sparse supervision, narrowing or even eliminating the gap with fully supervised methods.


\subsection{Ablation Study}

\begin{table}
\centering
\caption{Ablation study on the ACDC dataset. Various configurations of PCLMix are examined, incorporating supervised loss ($\ell_{sup}$), contrastive loss ($\ell_{ctr}$), heterogeneous consistency loss ($\ell_{het}$), and mixed consistency loss ($\ell_{mix}$). w/o tf represents the result without supervised loss of $\mathcal{F}_{t}$. }
\resizebox{\textwidth}{!}{
\begin{tabular}{*{13}{c}}
\toprule
\multirow{2}*{Versions} & \multirow{2}*{$\ell_{sup}$} & \multirow{2}*{$\ell_{ctr}$} & \multirow{2}*{$\ell_{het}$} & \multirow{2}*{$\ell_{mix}$} & \multicolumn{2}{c}{{\scriptsize RV }} & \multicolumn{2}{c}{{\scriptsize Myo }} & \multicolumn{2}{c}{{\scriptsize LV }} & \multicolumn{2}{c}{{\scriptsize Avg }} \\
\cmidrule(lr){6-7}\cmidrule(lr){8-9}\cmidrule(lr){10-11} \cmidrule(lr){12-13} & & & & & {\scriptsize Dice$\uparrow$ } & {\scriptsize HD95$\downarrow$ } & {\scriptsize Dice$\uparrow$ } & {\scriptsize HD95$\downarrow$ } & {\scriptsize Dice$\uparrow$ } & {\scriptsize HD95$\downarrow$ } & {\scriptsize Dice$\uparrow$ } & {\scriptsize HD95$\downarrow$ } \\
\midrule
\hash 1 & \checkmark & & & & 52.2 & 189.3 & 52.8 & 181.9 & 57.3 & 168.2 & 54.1 & 179.8 \\
\hash 2 & \checkmark & \checkmark & & & 58.8 & 147.0 & 54.7 & 144.2 & 61.4 & 136.9 & 58.3 & 142.7 \\
\hash 3 & \checkmark & & \checkmark & & 87.3 & 6.7 & 85.6 & 9.1 & 89.3 & 5.8 & 87.4 & 7.2 \\
\hash 4 & \checkmark & & \checkmark & \checkmark & 87.8 & \textbf{4.2} & \textbf{86.0} & 6.5 & \underline{91.7} & \textbf{3.1} & \underline{88.5} & \underline{4.6} \\
\hash 5 & \checkmark & \checkmark & \checkmark & & 86.6 & 6.4 & 85.5 & 7.2 & 89.6 & 5.3 & 87.9 & 6.3 \\
\hash 6 (w/o tf) & \checkmark  & \checkmark  & \checkmark & \checkmark & 
\underline{88.0} & 6.3 & 84.8 & \underline{4.0} & 91.5 & 4.4 & 88.1 & 4.9 \\
\hash 7 & \checkmark  & \checkmark  & \checkmark & \checkmark & \textbf{88.2} & \underline{5.3} & \underline{85.6} & \textbf{3.3} & \textbf{92.2} & \underline{3.7} & \textbf{88.7} & \textbf{4.1} \\
\bottomrule
\label{ablation}
\end{tabular}}
\end{table}

We conducted an ablation study to explore the impact of different loss components on experimental outcomes. Table \ref{ablation} summarizes results for various PCLMix configurations.

In version \hash 1, with only supervised loss, baseline performance is observed. Version \hash 7, incorporating all loss components, achieves peak performance with a Dice score of 88.7 and HD95 of 4.1. Comparison between versions \hash 1, \hash 3, and \hash 4 highlights the significant contribution of dual consistency regularization to improving model performance. Additionally, analyzing versions \hash 1, \hash 2, \hash 4, and \hash 7 reveals a limited impact of contrastive learning during poor model performance, but it proves beneficial under optimal conditions. We speculate that this is because, even with the guidance of uncertainty maps, contrastive learning still introduces noise pixels in situations of poor performance. The performance drop in version \hash 6, compared to \hash 7, emphasizes the Transformer's crucial role as an auxiliary decoder.

\subsection{Sensitivity Analysis}
\begin{figure}[htbp]
    \centering
    \subfigure[Sensitivity analysis of $\lambda_{t}$]{\includegraphics[width=0.48\columnwidth]{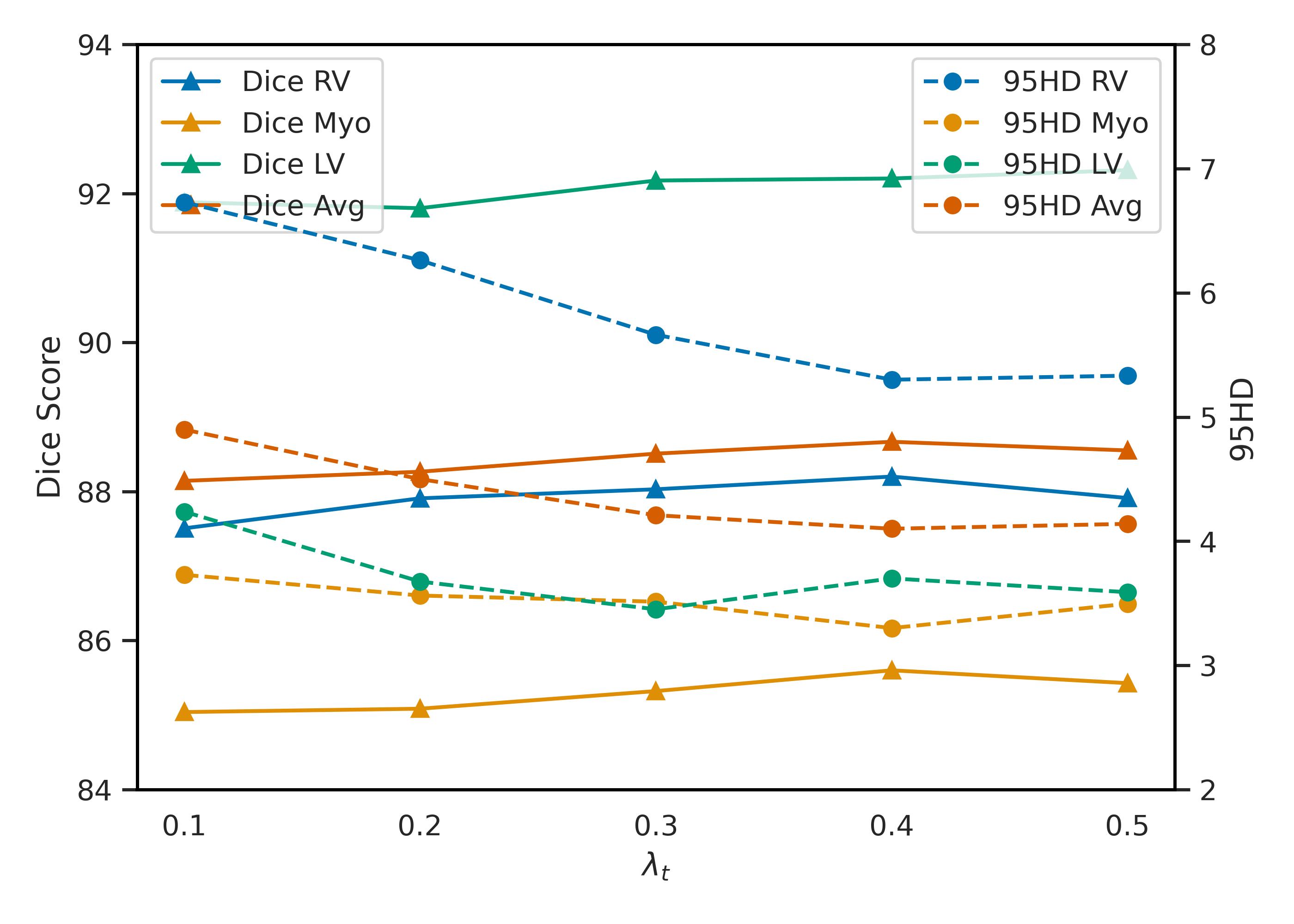}}\vspace{-10pt}
    \subfigure[Sensitivity analysis of $\lambda_{ctr}$]{\includegraphics[width=0.48\columnwidth]{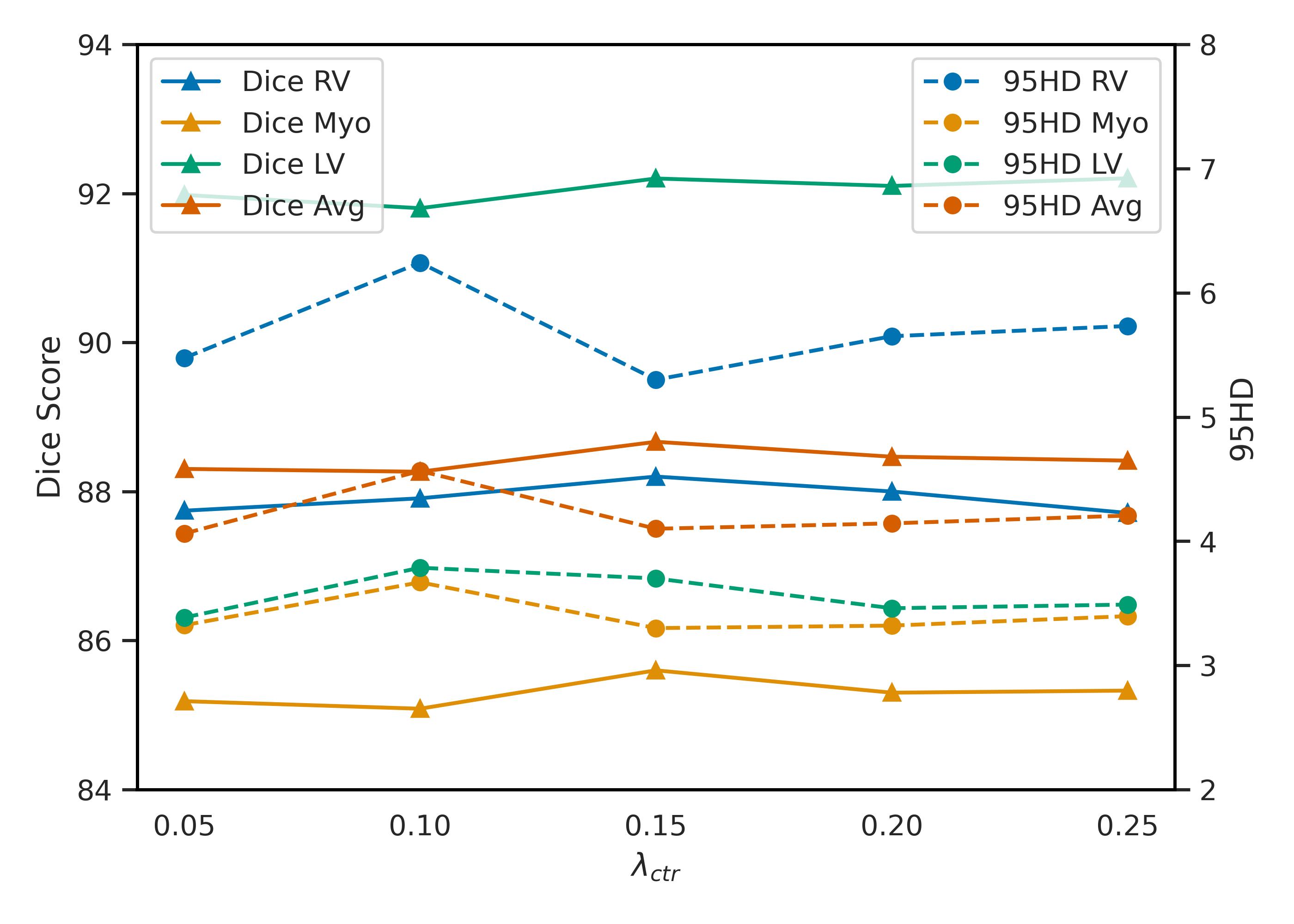}}
    \caption{Sensitivity analysis of transformer decoder weight ($\lambda_{t}$) and contrastive loss weight ($\lambda_{ctr}$).}
    \label{sensitivity}
\end{figure}

In the context of our dual-branch network architecture, the significance of the auxiliary branch cannot be overstated. Additionally, leveraging pixel-level contrastive learning guided by uncertainty proves to be instrumental in fortifying local predictions. To delve deeper into the intricate dynamics of our model, we conducted a detailed exploration of the sensitivity of two key parameters: $\lambda_{t}$, dictating the weight assigned to outputs of $\mathcal{F}_{t}$, and $\lambda_{ctr}$, determining the weight of the contrastive loss. We systematically varied $\lambda_{t}$ (exploring values like 0.1, 0.2, 0.3, 0.4, 0.5) while keeping other factors constant. Simultaneously, we examined the effects of changing $\lambda_{ctr}$ (testing values like 0.05, 0.10, 0.15, 0.2, 0.3). The results, shown in Figure \ref{sensitivity}, illustrate how segmentation outcomes evolve across RV, Myo, LV, and the average. Our findings reveal an interesting pattern: as $\lambda_{t}$ increases, experimental results steadily improve, reaching optimal performance at $\lambda_{t}=0.4$. However, further increments in $\lambda_{t}$ lead to a slight decline. Similarly, our experiments with $\lambda_{ctr}$ follow a similar trend, demonstrating optimal performance at $\lambda_{ctr}=0.15$.

\subsection{Visualization Analysis}
\begin{figure}[htbp]
    \centering
    \includegraphics[width=\linewidth]{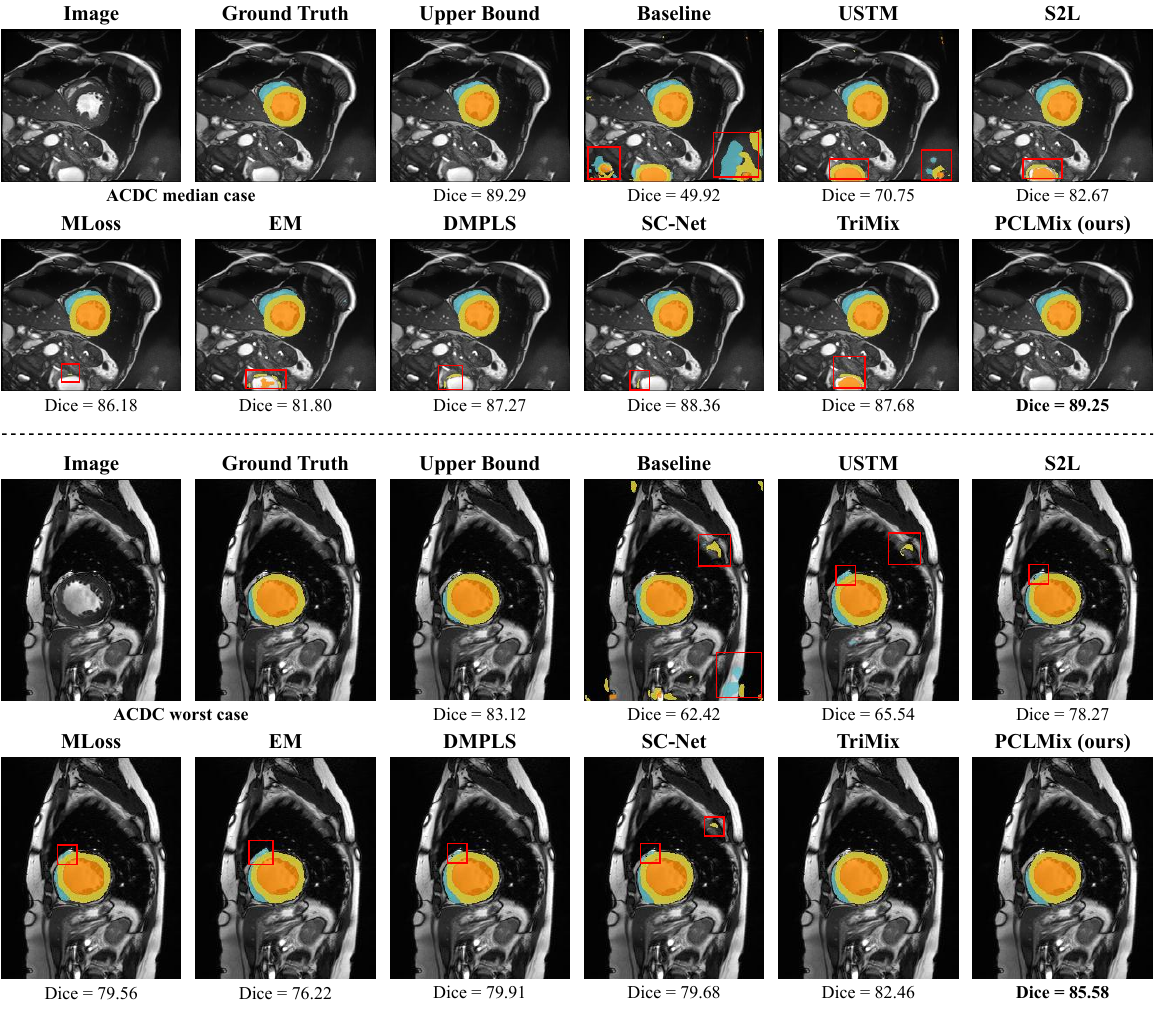}
    \caption{Qualitative comparison of different methods on ACDC dataset. The selected subjects were the median cases with regard to the Dice scores of the results of fully-supervised segmentation.}
    \label{visual_result}
\end{figure}

Figure \ref{visual_result} visualizes results on the median case selected using the upper bound method. It is observed that compared to SOTAs, PCLMix's predictions are closer to the ground truth, particularly in boundary regions and intricate local details, which once again suggests its superior segmentation performance. Due to the more condensed feature distributions, PCLMix mitigates false-positive predictions, as highlighted in the red box.

The experiments demonstrate exciting improvements of PCLMix in HD95, suggesting its superior segmentation performance, particularly in boundary regions, compared to the SOTAs, which can be further supported by Figure \ref{visual_result}. Furthermore, although PCLMix lags behind TriMix in Dice scores, it significantly reduces computational complexity and improves inference efficiency by discarding the auxiliary decoder branch during inference.

However, we also observed that introducing the proposed contrastive learning strategy may lead to further performance degradation, especially when the network's performance is poor. We attribute this to the inevitable introduction of more noisy pixels during anchor sampling, even with guidance from uncertainty maps. Lastly, due to the presence of the auxiliary decoder, PCLMix exhibits higher computational complexity during the training phase, limiting the exploration of ensemble methods due to resource constraints. We hope to address this limitation in future research endeavors.

\section{Conclusions}
In this work, we propose PCLMix for effective weakly supervised medical image segmentation using scribble annotations. Through dynamic mix augmentation, uncertainty-guided pixel-level contrastive learning, and dual consistency regularization, our approach alleviates two inherent challenges associated with sparse supervision: the lack of structural priors during training and the dispersed distribution of class features, enabling appropriate propagation of local supervision signals to the global scale. Comprehensive experiments on the public cardiac dataset ACDC demonstrate the comparability of our PCLMix with state-of-the-art scribble-supervised methods, significantly reducing or even eliminating the gap with fully supervised methods.

\subsubsection{Acknowledgments.} This work was supported by the National Natural Science Fund for Distinguished Young Scholar under Grant No.62025601, and the National Natural Science Foundation of China under Grant No.62376174.

%

\bibliographystyle{splncs04}
\bibliography{bibitem}

\end{document}